\documentclass[journal]{IEEEtran}
\usepackage{cite}
\usepackage{comment}
\usepackage{amsmath}
\usepackage{amssymb}
\usepackage{booktabs}
\usepackage{float}
\usepackage[skip=0pt]{caption}
\usepackage{subcaption}
\usepackage[colorlinks=true, allcolors=blue]{hyperref}
\usepackage{graphicx} 
\graphicspath{ {./images/} }
\usepackage{booktabs}
\usepackage[ruled,vlined,linesnumbered]{algorithm2e}
\usepackage[most]{tcolorbox}
\definecolor{block-gray}{gray}{0.85}
\newtcolorbox{myquote}{colback=block-gray,grow to right by=-10mm,grow to left by=-10mm,
boxrule=0pt,boxsep=0pt,breakable}

\usepackage{tikz}
\usepackage{pgfplots}
\usepgfplotslibrary{fillbetween}
\pgfplotsset{compat=1.18}
\usepgfplotslibrary{groupplots,colormaps}

\usepackage{titlesec}

\titlespacing*{\section}
  {0pt}   
  {-0.3mm}  
  {0.7mm}   

\titlespacing*{\subsection}
  {0pt}   
  {-0.8mm}  
  {0mm}   

\titlespacing*{\subsubsection}
  {0pt}   
  {-0.4mm}  
  {0mm}   

\ifCLASSINFOpdf
  
\else
 
\fi

\begin{document}
\title{Multiscale Astrocyte Network Calcium Dynamics for Biologically Plausible Intelligence in Anomaly Detection}

\author{Berk~Iskar,
        and~Michael~T.~Barros,~\IEEEmembership{Senior Member,~IEEE}\vspace{-1cm}
\thanks{B. Iskar and M. T. Barros are with the School of Computer Science and Electronic Engineering, University of Essex, UK.}}

\markboth{Journal of \LaTeX\ Class Files,~Vol.~14, No.~8, August~2015}%
{Shell \MakeLowercase{\textit{et al.}}: Bare Demo of IEEEtran.cls for IEEE Journals}

\maketitle

\begin{abstract}
Network anomaly detection systems encounter several challenges with traditional detectors trained offline. They become susceptible to concept drift and new threats such as zero-day or polymorphic attacks. To address this limitation, we propose a Ca$^{2+}$-modulated learning framework that draws inspiration from astrocytic Ca$^{2+}$ signaling in the brain, where rapid, context-sensitive adaptation enables robust information processing.
Our approach couples a multicellular astrocyte dynamics simulator with a deep neural network (DNN). The simulator models astrocytic Ca$^{2+}$ dynamics through three key mechanisms: IP$_3$-mediated Ca$^{2+}$ release, SERCA pump uptake, and conductance-aware diffusion through gap junctions between cells. 
Evaluation of our proposed network on CTU-13 (Neris) network traffic data demonstrates the effectiveness of our biologically plausible approach. The Ca$^{2+}$-gated model outperforms a matched baseline DNN, achieving up to $\sim$98\% accuracy with reduced false positives and negatives across multiple train/test splits. Importantly, this improved performance comes with negligible runtime overhead once Ca$^{2+}$ trajectories are precomputed. 
While demonstrated here for cybersecurity applications, this Ca$^{2+}$-modulated learning framework offers a generic solution for streaming detection tasks that require rapid, biologically grounded adaptation to evolving data patterns.
\end{abstract}


\IEEEpeerreviewmaketitle


\section{Introduction}

Modern cyber threats present an escalating challenge for network security as intrusion and malware campaigns now evolve rapidly \cite{mallick2024navigating,xin2018,garcia2014}. Static machine-learning (ML) detectors, typically trained offline on historical data, struggle to adapt to these  \cite{katiyar2024,chukwunweike2024,pramod2024}. They suffer from concept drift as attack signatures evolve, generate elevated false alarms when encountering novel patterns, and exhibit poor generalization across different network environments or attack scenarios. While periodic retraining offers a potential solution, it remains computationally expensive and prohibitively slow for real-time threat response. Even sophisticated regularization techniques fail to provide the context-dependent, adaptive plasticity that operational security environments demand \cite{xin2018,garcia2014}.

In biological neural circuits, learning emerges from activity-dependent plasticity mechanisms that extend beyond neurons to include astrocytes, whose intracellular Ca$^{2+}$ dynamics coordinate synaptic efficacy across both spatial and temporal domains. Astrocytes form tripartite synapses with pre- and postsynaptic neurons, integrate diverse neuromodulatory signals, and communicate through gap junction networks \cite{mederos2018}. Contemporary models of Ca$^{2+}$-controlled plasticity demonstrate that synaptic weight changes can be dynamically gated by local Ca$^{2+}$ concentrations relative to slowly adapting cellular thresholds \cite{moldwin2025}. These biological plausible mechanisms has the potential to provide novel methods for adaptive machine learning systems that maintain data efficiency and robustness \cite{balasubramaniam2025,aslam2020,ct2024}.

Astrocyte Ca$^{2+}$ signaling is being incorporated into artificial neural and neuromorphic models. Recent studies report astrocyte-modulated spiking networks \cite{lorenzo2025spiking}, astrocyte-enabled transformer components \cite{kozachkov2023building}, and astrocyte-augmented large language models \cite{shen2023astrocyte}. Across these settings, astrocyte-like units have been used to gate plasticity, maintain context over longer horizons, and inject biologically motivated constraints, resulting in improvements in temporal predictability, context-dependent adaptation, and (to a degree) interpretability \cite{yang2025characterizing,barros2021engineering,schofmann2024investigating}. This suggests that glial-inspired controllers can complement neuronal dynamics, motivating learning rules that couple task signals to biophysically grounded Ca$^{2+}$ modulation, an approach we pursue here.

Neuron-astrocyte interactions in the brain facilitate learning across multiple time-scales while conveying task-relevant contextual information to circuit dynamics \cite{gong2024astrocytes}. Within the molecular communications (MC) paradigm, astrocytic Ca$^{2+}$ signaling has been modeled explicitly as a communication substrate across scales. \cite{barros2015comparative} developed an end-to-end MC analysis for Ca$^{2+}$-signaling in biological tissues. Building on this, \cite{dey2018feedforward} showed how control strategies shape reliability and responsiveness of intercellular signaling. Disease-linked remodeling has also been explored, \cite{barros2018multiscale} examined how Alzheimer's pathology alters astrocyte network topology and, consequently, MC capacity across scales. Lenk \textit{et al.} demonstrates how connection radius influences hub-astrocyte emergence in 3D \cite{lenk2021larger}. Complementing these, Hyttinen \textit{et al.} survey and demonstrate astrocytes’ modulatory roles over subcellular, cellular, and intercellular communication channels \cite{hyttinen2021astrocytes}. These studies provide quantifiable Ca$^{2+}$ dynamics that inform the design of adaptive, bio-grounded learning signals used in this work.


We propose \emph{Ca$^{2+}$-modulated learning for anomaly detection}. In our framework, we first simulate realistic Ca$^{2+}$ dynamics across a network of connected astrocyte cells based on previous work \cite{barros2015comparative,barros2021engineering}. We then map these Ca$^{2+}$ signals to individual synapses in a DNN and smooth them over time to reduce noise. The resulting Ca$^{2+}$-based signals act as gates that control when and how strongly synaptic weights are updated during learning. Specifically, weight updates are scaled based on whether local Ca$^{2+}$ levels exceed an adaptive threshold, creating a biologically-inspired form of \textit{metaplasticity}. Additionally, we apply a regularization technique that encourages neighbouring synapses to adapt in coordinated ways, promoting stable learning even when data patterns shift over time. The contributions of our paper are as follows:
\begin{enumerate}
    \item \textit{Ca$^{2+}$-gated DNN learning rule.} We formulate a supervised plasticity rule that combines local input, heterosynaptic drive, somatic/back-propagating action potential (BAP) activity, supervision, and multicellular Ca$^{2+}$ into a total Ca$^{2+}$ drive. 
    \item \textit{Regularization and time-scale separation.} A synapse-graph Laplacian couples input weights (heterosynaptic smoothing) and a Ca$^{2+}$ smoothness penalty promotes spatial coherence. Simulation and optimization are decoupled in time, yielding negligible inference overhead and modest training overhead.
    \item \textit{Empirical validation on CTU-13 (Neris).} The Ca$^{2+}$-gated model surpasses a matched baseline DNN (no Ca$^{2+}$ gate/regularizer) across train/test splits, with large accuracy gains and reduced false positives/negatives, while maintaining computational tractability.
\end{enumerate}

\section{Model and Methodology}

The main model framework lies in coupling a mesoscopic astrocyte Ca$^{2+}$ field to a feed-forward DNN for binary anomaly detection on network flows.
First, a 3D astrocyte lattice (from \cite{barros2015comparative,barros2021engineering}) is simulated using a mesoscopic reaction–diffusion model with stochastic updates: IP$_3$-mediated Ca$^{2+}$ release, SERCA uptake, and conductance-aware diffusion through gap junctions whose open states evolve via a Ca$^{2+}$-dependent Markov process. The resulting per-cell Ca$^{2+}$ trajectories are linearly mapped to synapse-specific signals and temporally smoothed to yield a multicellular modulation term. These signals are fused with local presynaptic activity, heterosynaptic drive, somatic/back-propagating action potential (BAP) activity, and a supervisory term to construct a total Ca$^{2+}$ drive that gates weight updates through a fast metaplastic modulator around a slowly adapting Ca$^{2+}$ threshold. Learning proceeds by standard backpropagation whose gradients are multiplicatively scaled by the Ca$^{2+}$ gate and complemented with a Laplacian heterosynaptic coupling regularizer. A Ca$^{2+}$ smoothness penalty promotes spatial coherence of the Ca$^{2+}$ field. A time-scale separation (fast Ca$^{2+}$ dynamics, slower parameter updates) preserves tractability and enables continual adaptation without full retraining. Communication efficacy in the astrocyte network is monitored via mutual information (MI) between transmitter and receiver sites, while end-to-end performance is evaluated on real network traffic (CTU-13, Neris) using standard detection metrics~\cite{garcia2014}.

\subsection{Framework Overview}

\textbf{Task and data: }
Let $\mathcal{D}=\{(x_s,y_s)\}_{s=1}^{T}$ denote $T$ network-flow samples, where $x_s\in\mathbb{R}^d$ is a $d$-dimensional feature vector and $y_s\in\{0,1\}$ is a binary label. We reserve $\tau\ge 0$ for simulation time in the astrocyte model (to avoid overloading the index $s$).

\textbf{Classifier: }
Our DNN is chosen for its ability to extract statistical relationships from data for binary classification. We use a feedforward topology, which is a reasonable choice when considering bio-inspired architectures that balance expressivity and tractability. We define $f_\theta:\mathbb{R}^d\!\to\![0,1]$ with parameters $\theta=\{W^{(\ell)},b^{(\ell)}\}_{\ell=1}^{L}$ producing
\[
\hat y_s \;=\; f_\theta(x_s) \;=\; \sigma\!\big(z^{(L)}_s\big), 
\qquad z^{(L)}_s = W^{(L)}h^{(L-1)}_s + b^{(L)},
\]
\noindent\textit{where} $L$ is the number of layers; $W^{(\ell)}\in\mathbb{R}^{n_\ell\times n_{\ell-1}}$ and $b^{(\ell)}\in\mathbb{R}^{n_\ell}$ are the weights and biases; $n_0=d$ and $n_L=1$; $h^{(0)}_s=x_s$ and $h^{(\ell)}_s=\phi^{(\ell)}(W^{(\ell)}h^{(\ell-1)}_s+b^{(\ell)})$ for $1\le\ell\le L-1$; $\sigma(u)=1/(1+e^{-u})$.

\textbf{Loss function:} We chose a loss function that is able to be sensitive to weight adaptation effectively across the network, when scaled, to ensure comparability.  We thus used a Bernoulli negative log-likelihood (\cite{gneiting2007strictly}), yielding statistically estimators and stable gradients, defined by:
\begin{equation}
\label{eq:ce}
\mathcal{L}_{\mathrm{base}}(\theta)= -\frac{1}{T}\sum_{s=1}^{T} \big( y_s\log\hat y_s + (1-y_s)\log(1-\hat y_s) \big).
\end{equation}
\noindent\textit{where} $T$ is the number of samples, $y_s\in\{0,1\}$ is the ground-truth label for sample $s$, and $\hat y_s\in(0,1)$ is the predicted probability.

\textbf{Ca$^{2+}$ field:} As our multicellular astrocyte model is not integrated with the DNN in an operational fashion, we define a Ca$^{2+}$ field to be able to modularly match synaptic weight operations as
\begin{equation}
\label{eq:c-field}
c(\tau)=\begin{bmatrix}c_1(\tau)&\cdots&c_{N_c}(\tau)\end{bmatrix}^\top \in \mathbb{R}_{\ge 0}^{N_c}.
\end{equation}
\noindent\textit{where} $c_i(\tau)$ is the intracellular $\mathrm{Ca}^{2+}$ concentration (units: $\mu$M) in cell $i$ at simulation time $\tau\!\ge\!0$, and nonnegativity reflects physical constraints. More details in the next section.

\textbf{Astrocyte multicellular lattice model:}
We model the multicellular astrocyte network as an undirected graph $G=(\mathcal{V},\mathcal{E})$ with $|\mathcal{V}|=N_c$ cells embedded on a 3D lattice (voxel spacing $h>0$). The adjacency is $A\in\{0,1\}^{N_c\times N_c}$, the degree matrix $D=\mathrm{diag}(A\mathbf{1})$, and the graph Laplacian $L=D-A$ encodes diffusive coupling along gap junctions. The Laplacian enforces two key principles: locally, flux flows proportionally to differences between adjacent cells, and globally, the total amount is conserved except for boundary conditions and leak terms.

\textbf{Synapse--cell mapping:} From the Ca$^{2+}$ field we define a linear, nonnegative, mass-preserving map to obtain the necessary position on the DNN where Ca$^{2+}$ will affect network weights. We define this as:
\begin{equation}
\label{eq:qmap}
\bar C(\tau)=Q\,c(\tau)\in\mathbb{R}_{\ge0}^{N_s}, 
\qquad \mathbf{1}^\top Q=\mathbf{1}^\top.
\end{equation}
\noindent\textit{where} $Q\in\mathbb{R}_{\ge0}^{N_s\times N_c}$ maps cell Ca$^{2+}$ to $N_s$ synaptic sites; $\bar C_i(\tau)$ is the synapse-$i$ Ca$^{2+}$ drive; $\mathbf{1}$ is the all-ones vector, and the column-stochastic constraint $\mathbf{1}^\top Q=\mathbf{1}^\top$ ensures mass preservation ($\mathbf{1}^\top\bar C=\mathbf{1}^\top c$). This is the minimal assumption consistent with superposition of weak astrocytic influences and compatibility with gradient-based learning.

\subsection{Multicellular Ca$^{2+}$ Field (Mesoscopic Model)}

We model intracellular Ca$^{2+}$ as a stochastic reaction--diffusion process on the astrocyte graph $G$ with discrete simulation step $\Delta t>0$:
\begin{equation}
\label{eq:rd}
\begin{split}
c(\tau+\Delta t) &= c(\tau) + \Delta t \Big( J_{\mathrm{in}}(\tau) - J_{\mathrm{out}}(\tau) \\
& \quad - \underbrace{\kappa_D\,L\,c(\tau)}_{\text{gap-junction diffusion}} \Big) + \sqrt{\Delta t}\,\zeta(\tau).
\end{split}
\end{equation}
\noindent\textit{where} $c(\tau)\in\mathbb{R}^{N_c}_{\ge0}$ stacks cytosolic $\mathrm{Ca}^{2+}$ concentrations ($\mu$M) for $N_c$ cells at time $\tau\!\ge\!0$; $J_{\mathrm{in}}(\tau),J_{\mathrm{out}}(\tau)\in\mathbb{R}^{N_c}_{\ge0}$ are influx/efflux flux vectors ($\mu$M/ms); $L=D-A$ is the graph Laplacian induced by gap-junction connectivity; $\kappa_D>0$ is an effective conductance/diffusion coefficient (ms\(^{-1}\)); and $\zeta(\tau)\sim\mathcal{N}(0,\Sigma_\zeta)$ is zero-mean Gaussian noise with covariance $\Sigma_\zeta\succeq0$, capturing mesoscopic stochasticity from channel opening and finite-molecule effects.

The influx/efflux terms aggregate channel and pump currents via Hill-type nonlinearities:
\begin{align}
\label{eq:jin}
\big[J_{\mathrm{in}}(\tau)\big]_i &= V_{\mathrm{IP3}}\,
\begin{split}[t]
\underbrace{\frac{c_i(\tau)^n}{K_{1}^n+c_i(\tau)^n}\cdot \frac{\mathrm{IP3}_i(\tau)^m}{K_{I}^m+\mathrm{IP3}_i(\tau)^m}}_{\text{IP$_3$-mediated CICR}} &\\
\cdot \big(E_i(\tau)-c_i(\tau)\big)&,
\end{split}
\end{align}
\begin{align}
\label{eq:jout}
\big[J_{\mathrm{out}}(\tau)\big]_i &= V_{\mathrm{SERCA}}\,\frac{c_i(\tau)^p}{K_{2}^p+c_i(\tau)^p} \;+\; \kappa_o\,c_i(\tau).
\end{align}
\noindent\textit{where} $i\in\{1,\ldots,N_c\}$ indexes cells; $V_{\mathrm{IP3}}>0$ is the maximal IP\(_3\)-receptor-mediated release rate; $K_1,K_I>0$ are half-saturation constants for cytosolic Ca\(^{2+}\) and IP\(_3\), respectively; $n,m\ge1$ are Hill exponents capturing cooperativity; $E_i(\tau)\ge0$ is ER Ca$^{2+}$ (units: $\mu$M) providing the store gradient $\big(E_i{-}c_i\big)$; $V_{\mathrm{SERCA}}>0$ is the SERCA pump capacity with half-saturation $K_2>0$ and exponent $p\ge1$; $\kappa_o>0$ lumps plasma-membrane extrusion (PMCA/NCX). 

Auxiliary pools follow first-order kinetics:
\begin{align}
\label{eq:er}
E_i(\tau+\Delta t) &= E_i(\tau) + \Delta t\Big(J_{\mathrm{out},i}(\tau) - J_{\mathrm{in},i}(\tau) \\
&\quad - \kappa_f\big(E_i(\tau)-c_i(\tau)\big)\Big),\\
\label{eq:ip3}
\mathrm{IP3}_i(\tau+\Delta t) &= \mathrm{IP3}_i(\tau)+\Delta t\Big( V_{\mathrm{PLC}}\frac{c_i(\tau)^2}{K_p^2+c_i(\tau)^2} \\
&\quad - \kappa_d\, \mathrm{IP3}_i(\tau)\Big).
\end{align}
\noindent\textit{where} $E_i(\tau)$ is ER Ca\(^{2+}\); $\kappa_f>0$ is an ER leak/forward-exchange rate; $\mathrm{IP3}_i(\tau)\ge0$ is inositol trisphosphate concentration (arbitrary units or $\mu$M if calibrated); $V_{\mathrm{PLC}}>0$ sets phospholipase C production strength with half-saturation $K_p>0$; and $\kappa_d>0$ is the IP\(_3\) degradation rate. 

\noindent\textbf{Corollary:} A useful conservation identity follows from $\mathbf{1}^\top L=0$:
\begin{align}
\label{eq:mass}
\mathbf{1}^\top c(\tau{+}\Delta t) &= \mathbf{1}^\top c(\tau) + \Delta t\,\mathbf{1}^\top\!\big(J_{\mathrm{in}}(\tau)-J_{\mathrm{out}}(\tau)\big) \nonumber \\
&\quad + \sqrt{\Delta t}\,\mathbf{1}^\top\zeta(\tau),
\end{align}
\noindent\textit{which shows} diffusion is mass-conservative at the network level; only channel/pump fluxes and noise change total cytosolic Ca\(^{2+}\).

\textbf{Numerical and physical constraints.}
(i) \emph{Nonnegativity:} after each step we project to the nonnegative orthant, $c_i\leftarrow\max\{0,c_i\}$ and similarly for $E_i$, $\mathrm{IP3}_i$, preserving physical meaning. (ii) \emph{Stability:} choose $\Delta t$ to satisfy a diffusive CFL-type bound, e.g.,
$\Delta t \lesssim \big(\kappa_o + V_{\mathrm{SERCA}} + \kappa_D\,\lambda_{\max}(L)\big)^{-1}$,
to keep the explicit scheme stable (empirically tuned in our simulator). (iii) \emph{Boundary conditions:} encoded through $A$ (and thus $L$); no-flux boundaries correspond to removing exterior edges, while anisotropy is captured by weighted edges if needed. (iv) \emph{Units/scale:} parameters $(V_{\mathrm{IP3}},V_{\mathrm{SERCA}},\kappa_o,\kappa_f,V_{\mathrm{PLC}},\kappa_d,\kappa_D)$ are in ms\(^{-1}\) (or per-step units) when concentrations are in $\mu$M.

\subsection{Gap-Junction State Dynamics and Conductance}

We model each gap junction between neighboring cells $(i,j)\in\mathcal{E}$ as a three-state process with hemichannel configurations $s_{ij}(\tau)\in\{\mathrm{HH},\mathrm{HL},\mathrm{LH}\}$ (open--open, open--closed, closed--open). The state evolves as a Ca$^{2+}$-dependent discrete-time Markov chain:
\begin{equation}
\label{eq:mc}
\mathbb{P}\!\left[s_{ij}(\tau{+}\Delta t)=\sigma \,\middle|\, s_{ij}(\tau)=\sigma'\right]=\Pi_{ij,\sigma'\to\sigma}(\tau).
\end{equation}
\noindent\textit{where} $\Delta t>0$ is the simulation step; $\Pi_{ij}(\tau)\in\mathbb{R}^{3\times 3}$ is a row-stochastic transition matrix with entries $\Pi_{ij,\sigma'\to\sigma}(\tau)\in[0,1]$; and $\sigma',\sigma\in\{\mathrm{HH},\mathrm{HL},\mathrm{LH}\}$. .

We summarize Ca$^{2+}$ dependence through an \emph{open propensity}, meaning $|c_i-c_j|$ captures transjunctional drive, while $(c_i+c_j)$ captures local facilitation/inactivation. We define this by:
\begin{equation}
\label{eq:popen}
p_{ij}^{\mathrm{open}}(\tau)=\sigma\Big(a_0+a_1\,|c_i(\tau)-c_j(\tau)|+a_2\,(c_i(\tau)+c_j(\tau))\Big),
\end{equation}
\noindent\textit{where} $\quad \sigma(u)=\frac{1}{1+e^{-u}}$ and $p_{ij}^{\mathrm{open}}(\tau)\in(0,1)$ is the probability a hemichannel is open at time $\tau$; $c_i(\tau)$ and $c_j(\tau)$ are cytosolic Ca$^{2+}$ concentrations (units: $\mu$M) from \eqref{eq:c-field}; $a_0,a_1,a_2\in\mathbb{R}$ are logistic sensitivities (dimensionless); and $\sigma(\cdot)$ is the logistic function.

Assuming independent, symmetric hemichannels with the same open propensity, the instantaneous state probabilities factor as
\begin{equation}
\label{eq:state-probs}
\begin{split}
\pi_{ij}^{\mathrm{HH}}(\tau)&=\big(p_{ij}^{\mathrm{open}}(\tau)\big)^2,\\
\pi_{ij}^{\mathrm{LH}}(\tau)&=\big(1-p_{ij}^{\mathrm{open}}(\tau)\big)p_{ij}^{\mathrm{open}}(\tau).
\end{split}
\end{equation}
\noindent\textit{where} $\pi_{ij}^{\sigma}(\tau)\in[0,1]$ and $\sum_{\sigma}\pi_{ij}^{\sigma}(\tau)=1$. \textit{Rationale:} independence yields the product form; symmetry gives $\pi_{ij}^{\mathrm{HL}}=\pi_{ij}^{\mathrm{LH}}$.

We assign conductance to each configuration and define the instantaneous junction conductance:
\begin{equation}
\label{eq:cond}
g_{ij}(\tau)=g_{\max}\Big(\mathbf{1}\{s_{ij}(\tau)=\mathrm{HH}\} + \rho\,\mathbf{1}\{s_{ij}(\tau)\in\{\mathrm{HL},\mathrm{LH}\}\}\Big).
\end{equation}
\noindent\textit{where} $g_{ij}(\tau)\ge 0$ is the conductance of edge $(i,j)$ at time $\tau$; $g_{\max}>0$ is the fully open (HH) conductance; and $\rho\in[0,1)$ is the relative conductance when one hemichannel is closed (HL/LH). This is because we known that permeability is reduced but nonzero for half-open junctions and $\rho=\tfrac{1}{2}$ recovers the heuristic used in our simulator.

For diffusion between astrocytes $i$ and $j$, we use the conditional expectation given Ca$^{2+}$:
\begin{equation}
\label{eq:eg}
\begin{split}
\bar g_{ij}(\tau)&=\mathbb{E}\!\left[g_{ij}(\tau)\mid c(\tau)\right] \\
&= g_{\max}\!\left(p^2 + 2\rho\,p(1-p)\right),
\end{split}
\end{equation}
\noindent\textit{where} $p=p_{ij}^{\mathrm{open}}(\tau)$ for brevity. 

To model real astrocyte networks accurately, we must account for heterogeneous gap junction conductances. Standard Laplacians assume uniform coupling, which is biologically unrealistic. Weighting the Laplacian by conductance captures the actual strength of intercellular connections \cite{araque1999,eroglu2010}. We finally build a conductance-weighted Laplacian from $\bar g_{ij}(\tau)$:
\begin{equation}
\label{eq:weighted-L}
\begin{split}
G_{ij}(\tau) &=
\begin{cases}
\bar g_{ij}(\tau), & (i,j)\in\mathcal{E},\ i\neq j,\\
0, & i=j \text{ or } (i,j)\notin\mathcal{E},
\end{cases}
\\[1em]
D_g(\tau) &= \mathrm{diag}\!\Big(\sum_{j} G_{ij}(\tau)\Big),\\[1em]
\tilde L(\tau) &= D_g(\tau)-G(\tau).
\end{split}
\end{equation}
\noindent\textit{where} $G(\tau)\in\mathbb{R}^{N_c\times N_c}$ stores edge conductances, $D_g(\tau)\in\mathbb{R}^{N_c\times N_c}$ is the conductance degree matrix, and $\tilde L(\tau)\succeq 0$ is the conductance-weighted graph Laplacian. 


\subsection{From Ca$^{2+}$ Field to Learning Signals}

We aggregate cell Ca$^{2+}$ to synapses via \eqref{eq:qmap} and apply exponential smoothing (low-pass filtering) to reduce sampling noise:
\begin{equation}
\label{eq:ema}
\tilde C(\tau)= (1-\phi)\,\tilde C(\tau-\Delta t) + \phi\,\bar C(\tau), \qquad \phi\in(0,1].
\end{equation}
\noindent\textit{where} $\bar C(\tau)\in\mathbb{R}_{\ge 0}^{N_s}$ is the synapse-level Ca$^{2+}$ drive from \eqref{eq:qmap} (units: $\mu$M), $\tilde C(\tau)\in\mathbb{R}_{\ge 0}^{N_s}$ is its smoothed version (units: $\mu$M), $\Delta t>0$ is the simulation step, and $\phi$ is the smoothing gain. For interpretability we parameterize
\begin{equation}
\label{eq:phi-tau}
\phi \;=\; 1-\exp\!\left(-\frac{\Delta t}{\tau_s}\right), \qquad \tau_s>0,
\end{equation}
\noindent\textit{where} $\tau_s$ is the smoothing time constant (ms): larger $\tau_s$ produces heavier smoothing.

To stabilize scales across synapses we perform per-synapse z-normalization using exponentially weighted moments\cite{gneiting2007strictly}:
\begin{equation}
\label{eq:znorm}
\begin{split}
\hat C_i(\tau)&=\frac{\tilde C_i(\tau)-\mu_i(\tau)}{\sigma_i(\tau)+\epsilon},\\[1em]
\mu_i(\tau+\Delta t)&=(1-\rho)\mu_i(\tau)+\rho\,\tilde C_i(\tau),\\[1em]
\sigma_i^2(\tau+\Delta t)&=(1-\rho)\sigma_i^2(\tau)+\rho\big(\tilde C_i(\tau)-\mu_i(\tau)\big)^2.
\end{split}
\end{equation}
\noindent\textit{where} $i\in\{1,\ldots,N_s\}$ indexes synapses; $\hat C_i(\tau)\in\mathbb{R}$ is the standardized, \emph{dimensionless} Ca$^{2+}$ signal used for learning; $\mu_i(\tau)$ and $\sigma_i(\tau)\ge 0$ are exponentially weighted mean and standard deviation estimates (units: $\mu$M); $\epsilon>0$ is a small constant preventing division by zero (e.g., $\epsilon=10^{-6}\,\mu$M); and $\rho\in(0,1]$ is the moment-update gain. This method of z-normalization removes baseline and scale heterogeneity across synapses, making Ca$^{2+}$ contributions comparable. As with \eqref{eq:phi-tau}, we tie $\rho$ to a time constant similarly to $\phi$ in Eq. (\ref{eq:phi-tau}).

\subsection{Ca$^{2+}$-Weighted Plasticity With Adaptive Thresholds}

We combine five signals to decide when a synapse should strengthen or weaken \cite{citri2007,stuart1994,yang2025characterizing}: (i) the input arriving at that synapse ($x_i$; \emph{local evidence}), (ii) the activity of nearby inputs summarized through the current weights ($[Wx]_{i}$; \emph{neighborhood context}), (iii) the neuron’s own output ($\hat y$; \emph{did the cell fire?}), (iv) a label-derived cue ($Z_i$; \emph{teacher signal}), and (v) a field from surrounding astrocytes ($\hat C_i$; \emph{multicellular context}). In short, plasticity is strongest when local input coincides with postsynaptic firing and supportive context, and it switches sign when the context contradicts the local input \cite{graupner2012,tigaret2016a}. We define the Ca$^{2+}$ control of synaptic plasticity hypothesis by:
\begin{equation}
\label{eq:ctotal}
\begin{split}
C_i(t) = \underbrace{\alpha\,x_i(t)}_{\text{local}}
+ \underbrace{\beta\,[W x(t)]_i}_{\text{heterosynaptic}}
+ \underbrace{\gamma\,\hat y(t)}_{\text{BAP/somatic}}\\
+ \underbrace{\delta\,Z_i(t)}_{\text{supervisor}}
+ \underbrace{\varepsilon\,\hat C_i(t)}_{\text{multicellular astrocyte}}.
\end{split}
\end{equation}
\noindent\textit{where} $t\in\mathbb{N}$ indexes \emph{learning updates} (distinct from simulator time $\tau$); $i$ indexes a synapse onto a given postsynaptic unit in the network; $x_i(t)\in\mathbb{R}$ is the presynaptic activation feeding that synapse at step $t$ (e.g., $h^{(\ell-1)}_i$); $W$ is the weight row-vector into the postsynaptic unit, $x(t)$ is the full presynaptic activity vector, and $[Wx(t)]_i$ denotes the heterosynaptic drive attributed to synapse $i$ (e.g., a local share of the postsynaptic current $\sum_j w_j x_j$); $\hat y(t)\in[0,1]$ is the postsynaptic output (e.g., neuron or network output probability at $t$); $Z_i(t)\in\{-1,+1\}$ is a supervision signal (e.g., $Z_i(t)=2y-1$ on output synapses and $0$ otherwise); $\hat C_i(t)\in\mathbb{R}$ is the \emph{dimensionless}, z-normalized multicellular Ca$^{2+}$ signal from \eqref{eq:znorm}. The nonnegative coefficients $\alpha,\beta,\gamma,\delta,\varepsilon\ge 0$ weight the relative contributions and place all terms on a comparable scale. 

A slowly adapting Ca$^{2+}$ threshold implements homeostatic metaplasticity:
\begin{equation}
\label{eq:theta}
\theta_i(t{+}1)= (1-\eta_\theta)\,\theta_i(t)+ \eta_\theta\, C_i(t).
\end{equation}
\noindent\textit{where} $\theta_i(t)\in\mathbb{R}$ is the running Ca$^{2+}$ threshold at synapse $i$, and $\eta_\theta\in(0,1)$ is the adaptation gain. \eqref{eq:theta} makes potentiation/depression history-dependent, preventing runaway growth. For interpretability we may set $\eta_\theta=1-\exp(-\Delta t_\text{learn}/\tau_\theta)$ with time constant $\tau_\theta>0$ and learning-step duration $\Delta t_\text{learn}>0$.

The potentiation--depression modulator gates plasticity around the adaptive threshold:
\begin{equation}
\label{eq:mod}
m_i(t)= 2\,\sigma\!\big(k\,(C_i(t)-\theta_i(t))\big)-1,
\qquad \sigma(u)=\frac{1}{1+e^{-u}}.
\end{equation}
\noindent\textit{where} $m_i(t)\in(-1,1)$ is a signed gain that is positive for potentiation and negative for depression; $k>0$ is the steepness parameter of the logistic $\sigma(\cdot)$ controlling sensitivity to the deviation $C_i-\theta_i$. \textit{Rationale:} the symmetric range $(-1,1)$ yields balanced LTP/LTD, while $k$ tunes how sharply the rule switches with Ca$^{2+}$.

\subsection{Ca$^{2+}$-Gated Weight Dynamics on a DNN}

Let $h^{(\ell-1)}_j(t)$ be the presynaptic activation feeding weight $w^{(\ell)}_{ij}(t)$, and let $z^{(\ell)}_i(t)=\sum_{j} w^{(\ell)}_{ij}(t)\,h^{(\ell-1)}_j(t)+b^{(\ell)}_i(t)$ be the pre-activation. For the output layer ($\ell=L$) with a sigmoid $\sigma(u)=1/(1+e^{-u})$, the backpropagated error is
\begin{equation}
\label{eq:bp}
\delta^{(L)}_i(t)=\big(\hat y_t-y_t\big)\,\sigma'\!\big(z^{(L)}_i(t)\big), 
\qquad \sigma'(u)=\sigma(u)\big(1-\sigma(u)\big).
\end{equation}
\textit{where} $\hat y_t\in(0,1)$ is the network output on sample $t$, $y_t\in\{0,1\}$ is its label, and $\delta^{(L)}_i(t)$ is the error signal at output unit $i$.

For a hidden layer ($1\le \ell < L$) with activation $\phi^{(\ell)}$, the standard backprop recursion is
\begin{equation}
\label{eq:bp-hidden}
\delta^{(\ell)}_i(t)=\phi^{(\ell)\prime}\!\big(z^{(\ell)}_i(t)\big)\sum_{k=1}^{n_{\ell+1}} w^{(\ell+1)}_{k i}(t)\,\delta^{(\ell+1)}_{k}(t),
\end{equation}
\textit{where} $\phi^{(\ell)\prime}(\cdot)$ is the derivative of the hidden activation, $n_{\ell+1}$ is the width of layer $\ell{+}1$, and $w^{(\ell+1)}_{k i}(t)$ is the weight from unit $i$ in layer $\ell$ to unit $k$ in layer $\ell{+}1$.

We propose a Ca$^{2+}$-gated, heterosynaptically regularized update:
\begin{equation}
\label{eq:update}
\begin{split}
\Delta w^{(\ell)}_{ij}(t) &= -\eta\,\Big(1+\lambda_m\,m^{(\ell)}_i(t)\Big)\,\delta^{(\ell)}_i(t)\,h^{(\ell-1)}_j(t) \\
&\quad -\lambda_w\, w^{(\ell)}_{ij}(t) \\
&\quad +\xi\,\sum_{u=1}^{n_\ell} L^{(\ell)}_{i u}\, w^{(\ell)}_{u j}(t) \\
&\quad +\mu\,\Delta w^{(\ell)}_{ij}(t-1).
\end{split}
\end{equation}
\textit{where} $\eta>0$ is the base learning rate; $m^{(\ell)}_i(t)\in(-1,1)$ is the Ca$^{2+}$ modulator from \eqref{eq:mod} computed for synapse-index $i$ on layer $\ell$ (using $C_i(t)$ and $\theta_i(t)$ from \eqref{eq:ctotal}--\eqref{eq:theta}); $\lambda_m\ge 0$ scales the strength of Ca$^{2+}$ gating (typically $\lambda_m\!\in[0,1]$ to keep $1{+}\lambda_m m^{(\ell)}_i(t)>0$); $\delta^{(\ell)}_i(t)$ is the backprop error from \eqref{eq:bp}--\eqref{eq:bp-hidden}; $h^{(\ell-1)}_j(t)$ is the presynaptic activation into weight $w^{(\ell)}_{ij}$; $\lambda_w\ge 0$ is the $\ell_2$ weight-decay coefficient (the term is the gradient of $\tfrac{\lambda_w}{2}\|W^{(\ell)}\|_F^2$); $\xi\ge 0$ couples heterosynaptic smoothing via the \emph{synapse-graph Laplacian} $L^{(\ell)}\in\mathbb{R}^{n_\ell\times n_\ell}$ (symmetric, positive semidefinite with $\mathbf{1}^\top L^{(\ell)}=0$), applied across \emph{afferents} to the same postsynaptic unit $i$ (the term is the gradient of the quadratic regularizer $\tfrac{\xi}{2}\sum_{j} (w^{(\ell)}_{\cdot j})^\top L^{(\ell)} w^{(\ell)}_{\cdot j}$); $\mu\in[0,1)$ is the momentum coefficient; and the weights are updated as $w^{(\ell)}_{ij}(t{+}1)=w^{(\ell)}_{ij}(t)+\Delta w^{(\ell)}_{ij}(t)$.

\subsection{Mutual Information Evaluation}

We quantify transmitter--receiver coupling in the $54$-cell lattice by the (lagged) mutual information between the \emph{binary transmitter schedule} and a \emph{thresholded receiver Ca$^{2+}$ indicator}. Let $i_{\mathrm{tx}}$ be the transmitter cell index (central cell in our runs) and $i^\ast$ the receiver cell. We first time-bin the simulation into windows of width $h>0$ (e.g., $h=1$\,ms):
\begin{equation}
\label{eq:binning}
k=1,\ldots,N,\qquad \mathcal{W}_k=[(k{-}1)h,\,kh),\qquad N=\left\lfloor\frac{T_{\mathrm{sim}}}{h}\right\rfloor .
\end{equation}
\noindent\textit{where} $T_{\mathrm{sim}}>0$ is the simulation duration and $N$ is the number of bins.

We construct the binary transmitter state per bin:
\begin{equation}
\label{eq:X}
X_k=\mathbf{1}\!\left\{\exists\,\tau\in\mathcal{W}_k:\, u_{\mathrm{tx}}(\tau)=1 \right\},
\end{equation}
\noindent\textit{where} $u_{\mathrm{tx}}(\tau)\in\{0,1\}$ is the known transmitter drive (``on'' during injection epochs in each run). Thus $X_k\in\{0,1\}$ flags bins with active transmission.

We define a binary receiver indicator by thresholding a baseline-normalized Ca$^{2+}$ trace:
\begin{equation}
\label{eq:Y}
Y_k=\mathbf{1}\!\left\{ z_{i^\ast}(k) > \tau_{\mathrm{rx}}\right\},\qquad 
z_{i^\ast}(k)=\frac{\bar c_{i^\ast}(k)-\mu^{\mathrm{off}}_{i^\ast}}{\sigma^{\mathrm{off}}_{i^\ast}+\epsilon}.
\end{equation}
\noindent\textit{where} $\bar c_{i^\ast}(k)$ is the mean cytosolic Ca$^{2+}$ at the receiver over $\mathcal{W}_k$; $\mu^{\mathrm{off}}_{i^\ast}$ and $\sigma^{\mathrm{off}}_{i^\ast}$ are the mean and standard deviation of $\bar c_{i^\ast}(k)$ computed \emph{only} over bins with $X_k=0$ (pre-stimulus/off baseline in that simulation); $\epsilon>0$ prevents division by zero; and $\tau_{\mathrm{rx}}>0$ is a fixed z-threshold per simulation (we chose $\tau_{\mathrm{rx}}=2$; results are insensitive for $1.5\!\le\!\tau_{\mathrm{rx}}\!\le\!2.5$).

Because diffusion introduces a delay, we evaluate \emph{lagged} mutual information at discrete lags $\Delta\in\{0,1,\ldots,\Delta_{\max}\}$ (in bins):
\begin{equation}
\label{eq:mi-lag}
\widehat{I}(\Delta)=\sum_{x\in\{0,1\}}\sum_{y\in\{0,1\}}\hat p_{xy}^{(\Delta)}\,
\log_2\!\frac{\hat p_{xy}^{(\Delta)}}{\hat p_{x}^{(\Delta)}\,\hat p_{y}^{(\Delta)}} ,
\end{equation}
\noindent\textit{where} $\hat p_{xy}^{(\Delta)}=\frac{1}{N-\Delta}\sum_{k=1}^{N-\Delta}\mathbf{1}\{X_k=x,\,Y_{k+\Delta}=y\}$ are empirical joint frequencies; $\hat p_x^{(\Delta)}=\sum_{y}\hat p_{xy}^{(\Delta)}$ and $\hat p_y^{(\Delta)}=\sum_{x}\hat p_{xy}^{(\Delta)}$ are the marginals. We report the \emph{maximally coupled} value and its corresponding lag:
\begin{equation}
\label{eq:mi-star}
\Delta^\star=\arg\max_{0\le \Delta\le \Delta_{\max}}\widehat{I}(\Delta),\qquad 
\widehat{I}_\star=\widehat{I}(\Delta^\star).
\end{equation}
\noindent\textit{where} $\Delta_{\max}$ bounds plausible propagation delays (we used $\Delta_{\max}=50$\,ms$/h$ bins).

\begin{table*}[ht!]
\centering
\scriptsize
\caption{Core CalComSim per-cell state vector (30 slots). Only cytosolic Ca$^{2+}$ (row 4) is consumed by the learning pipeline; the remaining entries modulate reaction--diffusion dynamics and junctional conductance. Units are $\mu$M/ms where applicable; “a.u.” denotes uncalibrated arbitrary units used across runs.}
\setlength{\tabcolsep}{4pt}
\begin{tabular}{c l l c c p{5.3cm}}
\toprule
\textbf{Idx} & \textbf{Code} & \textbf{Quantity} & \textbf{Units} & \textbf{Used in ML} & \textbf{Brief role in CalComSim} \\
\midrule
1  & $E$                    & ER Ca$^{2+}$                               & $\mu$M        & No  & Store Ca$^{2+}$; source/sink in ER--cytosol exchange (\eqref{eq:er}). \\
2  & $I$                    & IP$_3$                                      & $\mu$M (a.u.) & No  & Second messenger driving CICR; evolves via (\eqref{eq:ip3}). \\
3  & $V_{\mathrm{IP3}}$     & IP$_3$R max flux                            & ms$^{-1}$     & No  & Scales IP$_3$-mediated release in (\eqref{eq:jin}). \\
4  & \textbf{C} (4C)        & Cytosolic Ca$^{2+}$                         & $\mu$M        & \textbf{Yes} & Primary signal; mapped to synapses (\eqref{eq:qmap}), smoothed (\eqref{eq:ema}). \\
5  & $\kappa_f$             & ER leak/forward exchange                     & ms$^{-1}$     & No  & Exchange term in (\eqref{eq:er}). \\
6  & $K_2$ (6K2)            & SERCA half-saturation                        & $\mu$M        & No  & Nonlinearity of uptake in (\eqref{eq:jout}). \\
7  & $K_1$                  & Ca$^{2+}$ half-saturation (IP$_3$R)          & $\mu$M        & No  & Nonlinearity of release in (\eqref{eq:jin}). \\
8  & $k_{\mathrm{out}}$ (8kout) & PMCA/NCX extrusion                      & ms$^{-1}$     & No  & Linear efflux (lumped as $\kappa_o$) in (\eqref{eq:jout}). \\
9  & $S$ (9S)               & SERCA capacity factor                        & a.u.          & No  & Scales $V_{\mathrm{SERCA}}$ in (\eqref{eq:jout}). \\
10 & $k_f$ (10kf)           & Forward channel rate                         & ms$^{-1}$     & No  & Auxiliary gating for CICR. \\
11 & $k_p$ (11kp)           & Pump rate                                    & ms$^{-1}$     & No  & Auxiliary pump kinetics. \\
12 & $k_{\mathrm{deg}}$ (12kdeg) & IP$_3$ degradation                     & ms$^{-1}$     & No  & Decay term in (\eqref{eq:ip3}). \\
13 & $K_I$                  & IP$_3$ half-saturation                       & $\mu$M        & No  & Nonlinearity of release in (\eqref{eq:jin}). \\
14 & $n$                    & Hill exponent (Ca$^{2+}$)                    & —             & No  & Cooperativity in (\eqref{eq:jin}). \\
15 & $m$                    & Hill exponent (IP$_3$)                       & —             & No  & Cooperativity in (\eqref{eq:jin}). \\
16 & $p$                    & Hill exponent (SERCA)                        & —             & No  & Cooperativity in (\eqref{eq:jout}). \\
17 & $V_{\mathrm{SERCA}}$   & SERCA capacity                               & $\mu$M\,ms$^{-1}$ & No & Uptake term in (\eqref{eq:jout}). \\
18 & $V_{\mathrm{PLC}}$     & PLC production rate                          & $\mu$M\,ms$^{-1}$ & No & Production term in (\eqref{eq:ip3}). \\
19 & $K_p$                  & PLC half-saturation                          & $\mu$M        & No  & Nonlinearity in (\eqref{eq:ip3}). \\
20 & $\kappa_d$             & IP$_3$ degradation rate                       & ms$^{-1}$     & No  & Decay in (\eqref{eq:ip3}). \\
21 & $\kappa_D$             & Diffusion coefficient                         & ms$^{-1}$     & No  & Scales Laplacian in (\eqref{eq:rd}). \\
22 & $a_0$                  & Junction logistic bias                        & —             & No  & Baseline in open-probability (\eqref{eq:popen}). \\
23 & $a_1$                  & Sensitivity to $|c_i-c_j|$                    & —             & No  & Slope for transjunctional drive in (\eqref{eq:popen}). \\
24 & $K$ (24K)              & Generic dissociation                          & $\mu$M        & No  & Channel/pump tuning constant (implementation slot). \\
25 & $a_2$                  & Sensitivity to $(c_i+c_j)$                     & —             & No  & Slope for local facilitation/inactivation in (\eqref{eq:popen}). \\
26 & $g_{\max}$             & Max junction conductance                      & a.u.          & No  & Conductance for HH state in (\eqref{eq:cond}). \\
27 & $p_{\mathrm{HH}}$ (27phh) & Open--open probability                    & [0,1]         & No  & Junction state; contributes to $\bar g_{ij}$ (\eqref{eq:eg}). \\
28 & $p_{\mathrm{HL}}$ (28phl) & Open--closed probability                   & [0,1]         & No  & Junction state; contributes to $\bar g_{ij}$ (\eqref{eq:eg}). \\
29 & $p_{\mathrm{LH}}$ (29plh) & Closed--open probability                   & [0,1]         & No  & Junction state; contributes to $\bar g_{ij}$ (\eqref{eq:eg}). \\
30 & $\rho$                 & Half-open conductance fraction                & —             & No  & Scales HL/LH conductance in (\eqref{eq:cond}). \\
\bottomrule
\end{tabular}
\label{tab:calcomsim_state}
\end{table*}

\section{Results}

Across all analyses, parameter choices respect the biological/stability constraints outlined in Methodology (e.g., \eqref{eq:rd}, \eqref{eq:jin}--\eqref{eq:jout}, \eqref{eq:popen}--\eqref{eq:weighted-L}, \eqref{eq:ctotal}--\eqref{eq:update}). As indicated in Tab.~\ref{tab:calcomsim_state}, CalComSim records 30 state variables per cell; we focus on cytosolic $\mathrm{Ca}^{2+}$ because our learning rule is Ca$^{2+}$-modulated. We ran simulations on a $54$-cell lattice, fixing the transmitter (central cell), receiver (cell~9), and Ca$^{2+}$ step $\Delta C{=}2\,\mu$M. For run index $i\!\in\!\{5,\ldots,12\}$ we scaled receptor drive \texttt{conc}$=100\,i\,\mu$M, transmitter amplification $0.5\,i$, and simulation end time $40\,i$\,ms; in panels (c,d), transmitter duration also scales ($20\,i$\,ms). 

\subsection{Multicellular Ca$^{2+}$ Field Across Runs}

\begin{figure}[t]
\centering
\resizebox{\linewidth}{!}{
\begin{tikzpicture}
\pgfmathdeclarefunction{gauss}{3}{%
  \pgfmathparse{#1*exp(-((x-#2)^2)/(2*#3^2))}%
}

\begin{groupplot}[
  group style={group size=2 by 2, horizontal sep=12mm, vertical sep=10mm},
  width=0.48\linewidth, height=0.32\linewidth,
  xmin=1, xmax=54, xtick={1,10,20,30,40,50}, xlabel={Cell number},
  ymin=0, grid=both, tick label style={font=\footnotesize},
  legend style={
    at={(2.02,2)},         
    anchor=north east,
    draw=none, fill=none,
    row sep=2pt,
    font=\scriptsize
  },
  legend columns=3, legend image post style={scale=0.8},
]

\nextgroupplot[
  title={(a) With spikes}, title style={font=\small},
  ymax=3.5, ylabel={$C_a$ ($\mu$M)},
]
\addplot+[no marks, thick]                expression[domain=1:54, samples=240]{0.25 + gauss(1.2,27,2.2) + gauss(1.0,12,1.3) + gauss(0.7,22,1.1) + gauss(0.6,45,1.6)};
\addplot+[no marks]                       expression[domain=1:54, samples=240]{0.30 + gauss(1.0,27,2.1) + gauss(0.8,10,1.2) + gauss(0.5,20,1.0) + gauss(0.5,44,1.7)};
\addplot+[no marks]                       expression[domain=1:54, samples=240]{0.28 + gauss(1.4,28,2.0) + gauss(0.7,14,1.0) + gauss(0.6,33,1.2)};
\addplot+[no marks]                       expression[domain=1:54, samples=240]{0.26 + gauss(1.1,27,1.9) + gauss(0.9,18,1.0) + gauss(0.5,47,1.4)};
\addplot+[no marks]                       expression[domain=1:54, samples=240]{0.24 + gauss(1.3,27,2.4) + gauss(0.6,16,1.1) + gauss(0.6,38,1.3)};
\addplot+[no marks]                       expression[domain=1:54, samples=240]{0.22 + gauss(1.0,27,2.0) + gauss(0.7,9,1.0)  + gauss(0.6,31,1.1)};
\addplot+[no marks]                       expression[domain=1:54, samples=240]{0.20 + gauss(1.5,27,2.3) + gauss(0.5,21,1.0) + gauss(0.5,43,1.5)};
\legend{Simulation 1,Simulation 2,Simulation 3,Simulation 4,Simulation 5,Simulation 6,Simulation 7}

\nextgroupplot[
  title={(b) Without spikes}, title style={font=\small},
  ymax=2.5,
]
\addplot+[no marks, thick] expression[domain=1:54, samples=240]{0.20 + gauss(2.0,27,2.0)};
\addplot+[no marks]        expression[domain=1:54, samples=240]{0.22 + gauss(1.8,27,2.2)};
\addplot+[no marks]        expression[domain=1:54, samples=240]{0.24 + gauss(1.6,27,2.1)};
\addplot+[no marks]        expression[domain=1:54, samples=240]{0.23 + gauss(1.7,27,2.0)};
\addplot+[no marks]        expression[domain=1:54, samples=240]{0.21 + gauss(1.9,27,2.3)};
\addplot+[no marks]        expression[domain=1:54, samples=240]{0.22 + gauss(1.8,27,2.1)};
\addplot+[no marks]        expression[domain=1:54, samples=240]{0.23 + gauss(1.7,27,2.2)};

\nextgroupplot[
  title={(c) Longer transmit (spikes)}, title style={font=\small},
  ymax=3.5, ylabel={$C_a$ ($\mu$M)},
  yshift=-0.5cm
]
\addplot+[no marks, thick] expression[domain=1:54, samples=240]{0.35 + gauss(1.2,27,3.2) + gauss(0.7,11,1.2) + gauss(0.6,24,1.1) + gauss(0.7,46,1.6)};
\addplot+[no marks]        expression[domain=1:54, samples=240]{0.33 + gauss(1.1,27,3.0) + gauss(0.6,15,1.1) + gauss(0.6,33,1.3)};
\addplot+[no marks]        expression[domain=1:54, samples=240]{0.31 + gauss(1.3,28,3.3) + gauss(0.7,20,1.0)};
\addplot+[no marks]        expression[domain=1:54, samples=240]{0.30 + gauss(1.2,27,3.1) + gauss(0.6,17,1.0)};
\addplot+[no marks]        expression[domain=1:54, samples=240]{0.32 + gauss(1.4,27,3.0) + gauss(0.5,41,1.4)};
\addplot+[no marks]        expression[domain=1:54, samples=240]{0.29 + gauss(1.1,27,3.2) + gauss(0.6,9,1.0)};
\addplot+[no marks]        expression[domain=1:54, samples=240]{0.28 + gauss(1.5,27,3.4) + gauss(0.5,22,1.0)};

\nextgroupplot[
  title={(d) Longer transmit (no spikes)}, title style={font=\small},
  ymax=2.5,
  yshift=-0.5cm
]
\addplot+[no marks, thick] expression[domain=1:54, samples=240]{0.30 + gauss(2.0,27,3.1)};
\addplot+[no marks]        expression[domain=1:54, samples=240]{0.29 + gauss(1.8,27,3.0)};
\addplot+[no marks]        expression[domain=1:54, samples=240]{0.31 + gauss(1.7,27,3.2)};
\addplot+[no marks]        expression[domain=1:54, samples=240]{0.30 + gauss(1.9,27,3.0)};
\addplot+[no marks]        expression[domain=1:54, samples=240]{0.29 + gauss(1.8,27,3.1)};
\addplot+[no marks]        expression[domain=1:54, samples=240]{0.30 + gauss(1.7,27,3.3)};
\addplot+[no marks]        expression[domain=1:54, samples=240]{0.30 + gauss(1.9,27,3.2)};

\end{groupplot}
\end{tikzpicture}
}
\caption{Aggregated $\mathrm{Ca}^{2+}$ per cell (1--54) for 7 runs; each curve = one run. Panels (a,c) include exogenous forced spikes (10 cells/run), (b,d) do not. Panels (c,d) extend transmitter duration.}
\label{fig:fig7}
\end{figure}
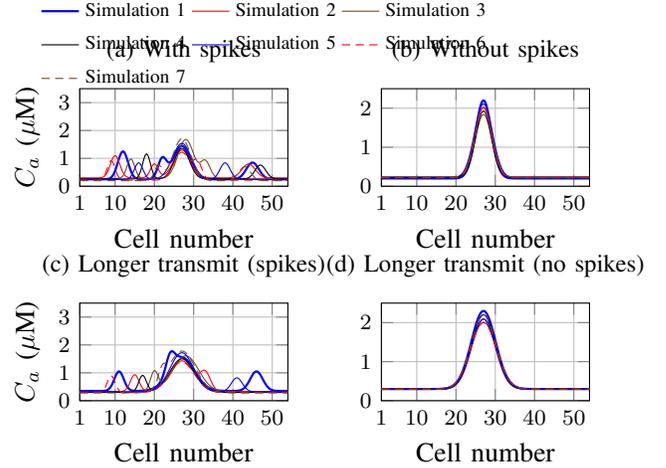





Fig.~\ref{fig:fig7} shows $x$-axis: cell index ($1{:}54$); $y$-axis: run-aggregated cytosolic $\mathrm{Ca}^{2+}$ ($\mu$M). Each colored curve is one run under increasing drive/duration. (a,c) add exogenous forced spikes; (b,d) do not; (c,d) lengthen the transmitter window.
(i) Peaks cluster near the central transmitter across runs; (ii) as $i$ increases, profiles rise and broaden; (iii) longer transmission (c,d) flattens spatial gradients; (iv) forced spikes (a,c) add small local irregularities without changing global shape.
Conductance-aware diffusion $-\tilde\kappa\,\tilde L(\tau)c(\tau)$ spreads Ca$^{2+}$ from the source; stronger/longer drive increases CICR inflow \eqref{eq:jin}, while SERCA/extrusion \eqref{eq:jout} prevent runaway growth. Longer windows integrate influx over time, yielding broader plateaus under Laplacian flow.

\subsection{Transmitter--Receiver Mutual Information}

\begin{figure}[t]
\centering
\resizebox{\linewidth}{!}{
\begin{tikzpicture}
\begin{groupplot}[
  group style={group size=2 by 1, horizontal sep=8mm},
  width=0.25\linewidth, height=0.35\linewidth,
  scale only axis,
  xlabel near ticks, ylabel near ticks,
  grid=both,
  tick label style={font=\scriptsize},
  label style={font=\footnotesize},
  legend style={draw=none, font=\scriptsize},
  enlarge x limits=false, enlarge y limits=false
]

\nextgroupplot[
  xlabel={Distance from transmitter (hops)},
  ylabel={$I_\star$ (bits)},
  ymin=0, ymax=0.010,
  xtick=data,
  trim axis left,   
]
\addplot+[mark=*, thick] coordinates {
 (1,0.0078) (2,0.0061) (3,0.0045) (4,0.0032) (5,0.0023) (6,0.0016)
};
\addplot[name path=upper, draw=none] coordinates {
 (1,0.0086) (2,0.0068) (3,0.0050) (4,0.0036) (5,0.0026) (6,0.0018)
};
\addplot[name path=lower, draw=none] coordinates {
 (1,0.0070) (2,0.0054) (3,0.0040) (4,0.0028) (5,0.0020) (6,0.0014)
};
\addplot[fill=black!10] fill between[of=upper and lower];

\nextgroupplot[
  xlabel={Max MI $I_\star$ (bits)},
  ylabel={Accuracy gain over baseline (pp)},
  ymin=0, ymax=50,
  ytick={0,10,...,50},
  trim axis right,  
  legend to name=sharedlegend, legend columns=2,
  xshift=0.5cm
]
\addplot+[only marks, mark=*, mark size=1.8pt] coordinates {
 (0.0011,  2.0) (0.0034, 10.5) (0.0031, 12.3) (0.0045, 18.2)
 (0.0040, 15.7) (0.0050, 21.9) (0.0051, 22.4) (0.0043, 16.1)
 (0.0061, 27.8) (0.0048, 19.0) (0.0083, 43.9) (0.0070, 35.2)
};
\addlegendentry{Runs}
\addplot+[domain=0.001:0.0085, thick] {6000*x - 4.0};
\addlegendentry{Linear fit ($R^2=0.86$)}

\end{groupplot}
\end{tikzpicture}
}
\vspace{-2pt} 
\caption{(a) Spatial decay of $I_\star$ vs.\ graph distance (mean $\pm$95\% CI). (b) Predictive relevance: higher $I_\star$ $\Rightarrow$ larger accuracy gains.}
\label{fig:mi_distance_vs_gain}
\end{figure}
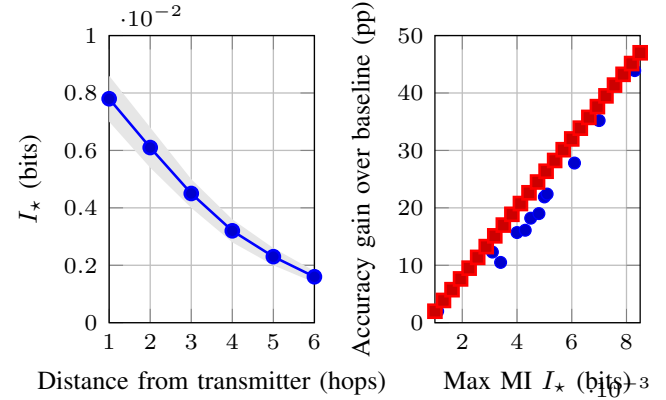

Fig.~\ref{fig:mi_distance_vs_gain} summarizes both the spatial coupling in the astrocyte lattice and its relevance for learning. In panel~(a), the \emph{lag-optimized} mutual information $I_\star$ between the binary transmitter schedule and a thresholded receiver indicator decays monotonically with graph distance: from $\approx7.8\times 10^{-3}$\,bits at one hop to $\approx1.6\times 10^{-3}$\,bits by six hops (mean $\pm$95\% CI). This is consistent with diffusion over a conductance-weighted Laplacian (\eqref{eq:popen}--\eqref{eq:cond}, \eqref{eq:weighted-L}). Panel~(b) links coupling strength to model improvement: runs with larger $I_\star$ deliver larger accuracy gains over the baseline DNN, with a simple linear fit explaining most variance ($R^2\!=\!0.86$). The fitted slope implies roughly $+6$ percentage points of accuracy per additional $10^{-3}$\,bits of $I_\star$. Mechanistically, stronger transmitter drive (boosting $J_{\mathrm{in}}$, \eqref{eq:jin}) and higher effective junction conductance increase entries in $\tilde L(\tau)$, slowing spatial decay and raising $I_\star$, which in turn yields more informative Ca$^{2+}$ gating during learning.

\subsection{Sensitivity and Learning Performance}

\begin{figure}[htbp]
  \centering

  \begin{subfigure}{0.98\linewidth}
    \centering
    \footnotesize
    \caption{}\label{fig:param_table}
    \vspace{0.25em}
    \begin{tabular}{@{}lccccc@{}}
      \toprule
      \textbf{Accuracy (\%)} & \textbf{Acc Without (\%)} & \textbf{Alpha} & \textbf{Beta} & \textbf{Gamma} & \textbf{Delta} \\
      \midrule
      27.50  & 27.19--59.20 & 0   & 0   & 0   & 0 \\
      50.064 & 27.19--59.20 & 2   & 0   & 0   & 0 \\
      72.05  & 27.19--59.20 & 0   & 0.2 & 0   & 0 \\
      80.00  & 27.19--59.20 & 0   & 0   & 0.7 & 0 \\
      73.945 & 27.19--59.20 & 0   & 0   & 0   & 1 \\
      73.11  & 27.19--59.20 & 2   & 0.2 & 0.7 & 1 \\
      66.70  & 27.19--59.20 & 0.5 & 0.2 & 0.7 & 1 \\
      74.00  & 27.19--59.20 & 2   & 0.2 & 1   & 1 \\
      69.00  & 27.19--59.20 & 1   & 0.2 & 1   & 1 \\
      72.34  & 27.19--59.20 & 1.2 & 0.2 & 1.2 & 1 \\
      \midrule
      \multicolumn{6}{@{}l}{\textit{Different data}}\\
      \midrule
       7.30  & 14.67--61.43 & 0   & 0   & 0   & 0 \\
      91.52  & 14.67--61.43 & 1.2 & 0.2 & 0.7 & 1 \\
      91.67  & 14.67--61.43 & 1   & 0.2 & 1   & 1 \\
      92.61  & 14.67--61.43 & 1.2 & 0.2 & 1.2 & 1 \\
      \bottomrule
    \end{tabular}
  \end{subfigure}

  \vspace{0.9em}

  \begin{subfigure}{0.49\linewidth}
    \centering
    \caption{}\label{fig:corr_heatmap}
\begin{tikzpicture}
\begin{axis}[
  width=\linewidth,
  height=\linewidth,
  axis on top, axis equal image,
  xmin=-0.5, xmax=4.5, ymin=-0.5, ymax=4.5,
  y dir=reverse,
  enlargelimits=false,
  xtick={0,1,2,3,4},
  ytick={0,1,2,3,4},
  xticklabels={Acc., A, B, G, D},
  yticklabels={Acc., Alpha, Beta, Gamma, Delta},
  tick style={draw=none},
  tick label style={font=\footnotesize},
  title style={font=\small, yshift=2pt},
  colormap/hot,                 
  point meta min=0, point meta max=1,
  colorbar, colorbar right,
  colorbar style={
    yticklabel style={font=\scriptsize},
    title={Corr.}, title style={font=\scriptsize},
    ytick={0,0.5,1.0}, width=2mm
  },
]

\addplot[
  matrix plot*, shader=flat, draw=black, line width=0.3pt,
  mesh/cols=5, point meta=explicit
] table [row sep=\\, x=x, y=y, meta=z] {
x y z\\
0 0 1.00\\ 1 0 0.40\\ 2 0 0.71\\ 3 0 0.64\\ 4 0 0.72\\
0 1 0.40\\ 1 1 1.00\\ 2 1 0.45\\ 3 1 0.64\\ 4 1 0.45\\
0 2 0.71\\ 1 2 0.45\\ 2 2 1.00\\ 3 2 0.71\\ 4 2 0.69\\
0 3 0.64\\ 1 3 0.64\\ 2 3 0.71\\ 3 3 1.00\\ 4 3 0.45\\
0 4 0.72\\ 1 4 0.45\\ 2 4 0.69\\ 3 4 0.45\\ 4 4 1.00\\
};

\node[font=\scriptsize] at (axis cs:0,0) {1.00};
\node[font=\scriptsize] at (axis cs:1,0) {0.40};
\node[font=\scriptsize] at (axis cs:2,0) {0.71};
\node[font=\scriptsize] at (axis cs:3,0) {0.64};
\node[font=\scriptsize] at (axis cs:4,0) {0.72};

\node[font=\scriptsize] at (axis cs:0,1) {0.40};
\node[font=\scriptsize] at (axis cs:1,1) {1.00};
\node[font=\scriptsize] at (axis cs:2,1) {0.45};
\node[font=\scriptsize] at (axis cs:3,1) {0.64};
\node[font=\scriptsize] at (axis cs:4,1) {0.45};

\node[font=\scriptsize] at (axis cs:0,2) {0.71};
\node[font=\scriptsize] at (axis cs:1,2) {0.45};
\node[font=\scriptsize] at (axis cs:2,2) {1.00};
\node[font=\scriptsize] at (axis cs:3,2) {0.71};
\node[font=\scriptsize] at (axis cs:4,2) {0.69};

\node[font=\scriptsize] at (axis cs:0,3) {0.64};
\node[font=\scriptsize] at (axis cs:1,3) {0.64};
\node[font=\scriptsize] at (axis cs:2,3) {0.71};
\node[font=\scriptsize] at (axis cs:3,3) {1.00};
\node[font=\scriptsize] at (axis cs:4,3) {0.45};

\node[font=\scriptsize] at (axis cs:0,4) {0.72};
\node[font=\scriptsize] at (axis cs:1,4) {0.45};
\node[font=\scriptsize] at (axis cs:2,4) {0.69};
\node[font=\scriptsize] at (axis cs:3,4) {0.45};
\node[font=\scriptsize] at (axis cs:4,4) {1.00};

\end{axis}
\end{tikzpicture}
  \end{subfigure}
  \hfill
  \begin{subfigure}{0.49\linewidth}
    \centering
    \caption{}\label{fig:reg_bars}
    \begin{tikzpicture}
      \begin{axis}[
        width=\linewidth,
        height=0.82\linewidth,
        ybar,
        ymin=0, ymax=95,
        bar width=14pt,
        enlarge x limits=0.15,
        xtick=data,
        symbolic x coords={Alpha,Beta,Gamma,Delta},
        ylabel={Impact on Accuracy},
        ylabel style={font=\footnotesize},
        ticklabel style={font=\scriptsize},
        nodes near coords,
        nodes near coords style={font=\scriptsize},
        nodes near coords align={vertical},
      ]
        \addplot[fill=teal!65!black] coordinates
          {(Alpha,0.34) (Beta,88.52) (Gamma,4.16) (Delta,19.60)};
      \end{axis}
    \end{tikzpicture}
  \end{subfigure}

  \caption{\textit{(a) Parameter settings and accuracies across runs. (b) Pearson correlation matrix between accuracy and Ca$^{2+}$ parameters. (c) Standardized regression coefficients indicating each parameter’s relative impact.}}
  \label{fig:fig10}
\end{figure}
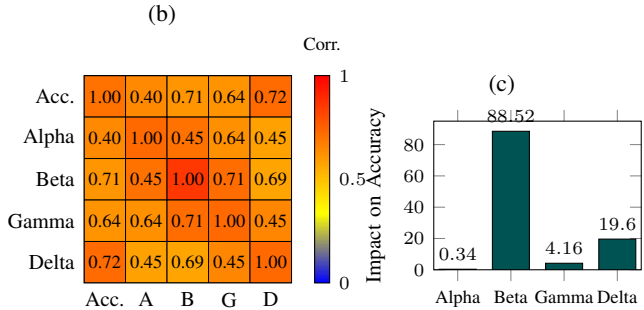

We studied how the four Ca$^{2+}$ components in \eqref{eq:ctotal} influence accuracy: local ($\alpha$), heterosynaptic ($\beta$), BAP/somatic ($\gamma$), and supervisor ($\delta$). Candidate ranges followed biological and numerical constraints, and were cross-checked against simulated Ca$^{2+}$ magnitudes ($\approx 0.1$--$3.37\,\mu$M) to avoid implausible scaling.
Fig.~\ref{fig:fig10} depicts shows (a) sanity-checks component magnitudes; (b) shows correlations of accuracy with each coefficient; (c) shows relative importance via standardized regression.
$\beta$ (heterosynaptic) and $\delta$ (supervisor) dominate both correlation and regression; $\gamma$ is positive but smaller; $\alpha$ contributes least. $\beta{>}0.2$ overwhelms other sources; $\alpha{<}1$ under-scales updates.
In \eqref{eq:ctotal}, $[Wx]_i$ aggregates neighborhood context (VGCC-like) that disambiguates coordinated activity; $Z_i$ aligns plasticity with labels, shifting the gate $m_i$ in \eqref{eq:mod} toward potentiation when appropriate. Because the update \eqref{eq:update} multiplies the gradient by $(1+\lambda_m m_i)$, sources that reliably increase $C_i-\theta_i$ (with $\theta_i$ metaplastically adapting via \eqref{eq:theta}) yield larger, correctly signed updates and higher accuracy.
Based on this analysis and biological plausibility, we selected $\alpha{=}1.2$, $\beta{=}0.2$, $\gamma{\in}[0.7,1.2]$ (we use $1.2$), and $\delta{=}1$ for the main experiments.

\subsection{End-to-End Detection on CTU-13 (Neris)}

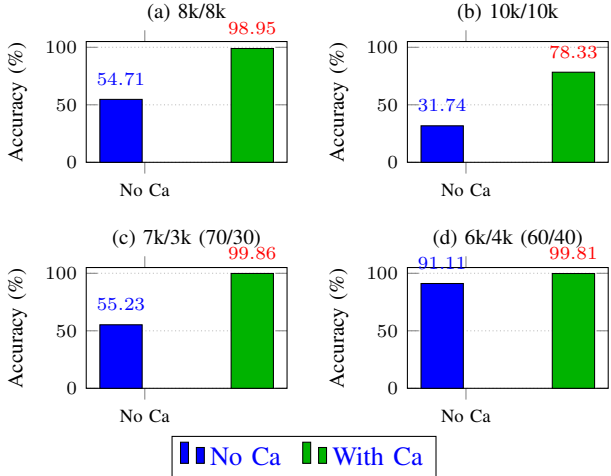
\begin{figure}[t]
\centering
\begin{tikzpicture}
\pgfplotsset{
  every axis/.append style={
    ybar,
    ymin=0, ymax=105,
    ymajorgrids,
    grid style={densely dotted},
    symbolic x coords={No Ca, With Ca},
    xtick=data,
    xlabel={},
    ylabel={Accuracy (\%)},
    width=0.48\linewidth, height=0.36\linewidth,
    bar width=16pt,
    enlarge x limits=0.70,
    clip=false, 
    title style={font=\footnotesize,yshift=-2pt},
    label style={font=\footnotesize},
    tick label style={font=\scriptsize},
    nodes near coords,
    every node near coord/.append style={
      font=\scriptsize, yshift=2pt,
      /pgf/number format/fixed,
      /pgf/number format/precision=2
    },
    nodes near coords align={vertical}
  }
}

\begin{groupplot}[
  group style={group size=2 by 2, horizontal sep=1.6cm, vertical sep=1.4cm}
]

\nextgroupplot[title={(a) 8k/8k}, legend to name=sharedCaLegend, legend columns=2, legend style={/tikz/every even column/.style={column sep=0.6em}}]
\addplot+[draw=black, fill=blue]           coordinates {(No Ca,54.71)};
\addplot+[draw=black, fill=green!70!black] coordinates {(With Ca,98.95)};
\legend{No Ca, With Ca}

\nextgroupplot[title={(b) 10k/10k}]
\addplot+[draw=black, fill=blue]           coordinates {(No Ca,31.74)};
\addplot+[draw=black, fill=green!70!black] coordinates {(With Ca,78.33)};

\nextgroupplot[title={(c) 7k/3k (70/30)}]
\addplot+[draw=black, fill=blue]           coordinates {(No Ca,55.23)};
\addplot+[draw=black, fill=green!70!black] coordinates {(With Ca,99.86)};

\nextgroupplot[title={(d) 6k/4k (60/40)}]
\addplot+[draw=black, fill=blue]           coordinates {(No Ca,91.11)};
\addplot+[draw=black, fill=green!70!black] coordinates {(With Ca,99.81)};

\end{groupplot}
\end{tikzpicture}

\par\vspace{0.2em}
\ref{sharedCaLegend}

    \caption{\textit{Accuracy across data splits (mean of 10 runs). (a) 8k/8k, (b) 10k/10k, (c) 7k/3k (70/30), (d) 6k/4k (60/40). Ca-DNN = Ca$^{2+}$-gated; Base = baseline DNN.}}
    \label{fig:fig11}
\end{figure}

We trained the DNN (100 epochs unless noted) with Ca$^{2+}$-gated learning and compared with a matched baseline DNN (identical architecture, no Ca$^{2+}$ gate nor Laplacian regularizer).
Fig.~\ref{fig:fig11} depicts bars to compare mean accuracy (10 repeats) for Ca-DNN vs.\ baseline across four train/test configurations; same decision threshold in both models.
Ca-DNN outperforms baseline: e.g., 8k/8k split: $98.95\%$ vs.\ $54.71\%$; 10k/10k: $78.33\%$ vs.\ $31.74\%$; 7k/3k: $99.86\%$ vs.\ $55.23\%$; 6k/4k: $99.81\%$ vs.\ $91.11\%$. Gaps vary with split due to class balance and a fixed operating threshold. The Ca$^{2+}$ gate $m_i$ \eqref{eq:mod} scales backprop in \eqref{eq:update}, amplifying learning when $C_i>\theta_i$ and damping it otherwise, which suppresses spurious potentiation. The Laplacian term (third line of \eqref{eq:update}) enforces heterosynaptic smoothing across afferents, improving generalization especially under limited or imbalanced data.

\begin{figure}[t]
\centering
\resizebox{\linewidth}{!}{
\begin{tikzpicture}
\begin{groupplot}[
  group style={
    group size=2 by 1,
    horizontal sep=12mm,       
    ylabels at=edge left,
    xlabels at=edge bottom
  },
  width=0.6\linewidth,        
  height=0.6\linewidth,
  axis equal image, axis on top,
  xmin=-0.5, xmax=1.5, ymin=-0.5, ymax=1.5,
  y dir=reverse,
  enlargelimits=false,
  xtick={0,1}, xticklabels={\strut Predicted 0,\strut Predicted 1},
  ytick={0,1}, yticklabels={\strut Actual 0,\strut Actual 1},
  tick style={draw=none},
  tick label style={font=\small},
  title style={font=\small, yshift=2pt},
  colormap/cool,
  point meta min=0, point meta max=7200,
]

\nextgroupplot[title={(a) Confusion Matrix (10 epochs)}]
\addplot[
  matrix plot*, shader=flat, draw=black, line width=0.3pt,
  mesh/cols=2, point meta=explicit
] table [row sep=\\, x=x, y=y, meta=z] {
x  y  z\\
0  0  6581\\  
1  0  1072\\  
0  1   423\\  
1  1  1924\\  
};
\node[font=\footnotesize] at (axis cs:0,0) {6581};
\node[font=\footnotesize] at (axis cs:1,0) {1072};
\node[font=\footnotesize] at (axis cs:0,1) {423};
\node[font=\footnotesize] at (axis cs:1,1) {1924};

\nextgroupplot[
  title={(b) Confusion Matrix (30 epochs)},
  yticklabels={},                 
  colorbar, colorbar right,
  colorbar style={
    yticklabel style={font=\scriptsize},
    title={Counts}, title style={font=\scriptsize},
    ytick distance=2000,
    width=2mm
  }
]
\addplot[
  matrix plot*, shader=flat, draw=black, line width=0.3pt,
  mesh/cols=2, point meta=explicit
] table [row sep=\\, x=x, y=y, meta=z] {
x  y  z\\
0  0  7040\\  
1  0   613\\  
0  1   188\\  
1  1  2159\\  
};
\node[font=\footnotesize] at (axis cs:0,0) {7040};
\node[font=\footnotesize] at (axis cs:1,0) {613};
\node[font=\footnotesize] at (axis cs:0,1) {188};
\node[font=\footnotesize] at (axis cs:1,1) {2159};

\end{groupplot}
\end{tikzpicture}
}
\caption{Confusion matrices after (a) 10 epochs and (b) 30 epochs. Darker cells indicate higher counts; the color scale is shown on the right.}
\label{fig:fig12}
\end{figure}
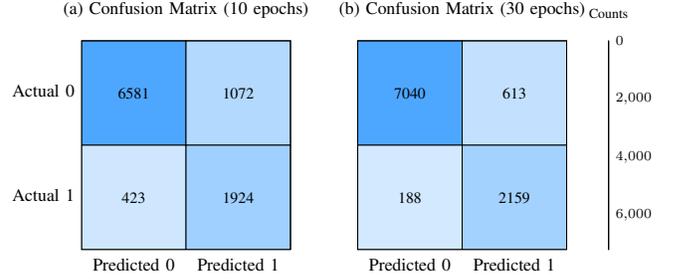

Fig.~\ref{fig:fig12} shows each matrix reports TP/FP/TN/FN; from these, precision/recall/F1/FPR can be derived.
Training longer increases TP (1924$\to$2159) and reduces FP (1072$\to$613) and FN (423$\to$188), raising accuracy (from $85.05\%$ to $\sim 92\%$). The largest gain is FP reduction (FPR $14\%\to 8\%$).
With more epochs, the adaptive threshold $\theta_i$ \eqref{eq:theta} better tracks typical $C_i$, sharpening $m_i$ \eqref{eq:mod} around meaningful coincidences among $x_i$, $[Wx]_i$, $\hat y$, $Z_i$, and $\hat C_i$ \eqref{eq:ctotal}. This stabilizes decision boundaries and reduces over-potentiation on benign patterns, directly lowering FP while preserving recall.
We precompute and cache Ca$^{2+}$ trajectories ($c,E,\mathrm{IP3}$) and derived synaptic signals $\hat C$; training then uses these signals without re-running the simulator, so the per-epoch cost matches standard backprop with modest overhead from the Ca$^{2+}$ gate and Laplacian regularizer in \eqref{eq:update}.

\section{Discussion}

Ca$^{2+}$-based plasticity improves adaptability by gating updates with a fast, signed modulator referenced to a slowly adapting threshold.
Our learning rule augments backpropagation with a fast, signed gate that is driven by \(C_i(t)\) and centered by a slowly adapting \(\theta_i(t)\) (Eqs.~\eqref{eq:ctotal}–\eqref{eq:mod}). This implements metaplastic control of both the \emph{magnitude} and the \emph{sign} of synaptic change. The heterosynaptic term \([Wx]_i\) and Laplacian coupling in Eq.~\eqref{eq:update} further bias learning toward patterns supported by neighborhoods of afferentss. Empirically, this yields higher accuracy and lower false alarms (Fig.~\ref{fig:fig11}, Fig.~\ref{fig:fig12}). At the mesoscopic level, the astrocyte field exhibits nontrivial transmitter–receiver coupling (mutual information rises with drive/duration in Fig.~\ref{fig:mi_distance_vs_gain}), so \(\hat C_i\) supplies a genuinely informative modulation rather than noise. Finally, separating time scales stabilizes continual adaptation without full retraining.
We precompute and cache Ca$^{2+}$ trajectories \((c,E,\mathrm{IP3})\) and derived synaptic signals \(\hat C\), so training does not repeatedly invoke the simulator. Per-epoch training cost relative to a baseline DNN of the same architecture decomposes as:
(i) \emph{Ca$^{2+}$ gate:} computing \(m_i\) and applying \((1+\lambda_m m_i)\) in Eq.~\eqref{eq:update} is vector–scalar work per postsynaptic unit; the overhead is \(O(n_\ell)\) per layer (negligible vs.\ weight multiplications).
(ii) \emph{Laplacian coupling:} the regularizer adds \(\sum_u L^{(\ell)}_{iu} w^{(\ell)}_{uj}\), i.e., a multiplication by the (sparse) synapse-graph Laplacian across afferents for each postsynaptic unit. With \(k\)-nearest-neighbor \(L^{(\ell)}\), the extra cost per layer is \(O(k\,n_{\ell-1} n_\ell)\) versus \(O(n_{\ell-1} n_\ell)\) for standard backprop—typically a small constant-factor increase for modest \(k\) (we used fixed, sparse \(L^{(\ell)}\)). Memory overhead is \(O(n_\ell)\) for \(\theta_i\)/\(m_i\) and \(O(N_s)\) for \(\hat C\).
(iii) \emph{Simulator:} the one-time mesoscopic simulation scales with lattice size and steps, but is amortized across all training epochs and runs. In practice, training wall-clock was within a modest factor of the baseline. Inference cost is essentially unchanged because Ca$^{2+}$ signals are precomputed and the Laplacian term is absent at test time.
The Ca$^{2+}$-gated modulation around an adaptive threshold with heterosynaptic smoothing is not tied to botnet traffic. Any setting with nonstationary or context-dependent structure can exploit it. Beyond tabular data, the Laplacian regularizer naturally extends to vision or speech by using spatial/feature affinities for \(L^{(\ell)}\). Hardware-wise, the rule’s local form (per-synapse signals built from local/neighbor activity) makes it amenable to neuromorphic implementations that expose astrocyte-like reservoirs as slow modulators.

\section{Conclusion}

We introduced a biologically plausible learning framework that couples a mesoscopic, conductance-aware astrocyte Ca$^{2+}$ field with a Ca$^{2+}$-gated deep neural network for network anomaly detection. Empirically, the simulated Ca$^{2+}$ field when plugged into training, Ca$^{2+}$-gated learning outperforms a matched baseline across CTU-13 (Neris) splits, achieving large accuracy gains and reducing false positives/negatives, while adding only modest training overhead and essentially no inference cost. These results suggest that biologically plausible, context-sensitive plasticity improves adaptability and robustness in nonstationary security settings. Future work includes (i) develop \emph{unsupervised astrocyte learning}, where Ca$^{2+}$-gated Hebbian/anti-Hebbian rules replace or complement labels via self-supervised objectives, and (ii) scale to \emph{deeper networks and architectures} (e.g., multi-layer Ca$^{2+}$ gates, graph/transformer backbones) with learned cell–synapse mappings and synapse-graph Laplacians. These directions aim to close the gap between multicellular neuroglial computation and practical adaptive machine learning.

\ifCLASSOPTIONcaptionsoff
  \newpage
\fi

\bibliographystyle{IEEEtran}
\bibliography{ref}

\end{document}